\pdfoutput=1

\documentclass[11pt]{article}

\usepackage[]{acl}

\usepackage[title]{appendix}
\usepackage{times}
\usepackage{latexsym}
\usepackage{hyperref}
\usepackage[T1]{fontenc}
\usepackage[utf8]{inputenc}
\usepackage{microtype}
\usepackage{svg}
\usepackage{caption}
\usepackage{subcaption}
\usepackage{bm}
\usepackage{multirow}
\usepackage{lipsum}
\usepackage{amsmath}
\usepackage{amssymb}
\DeclareMathOperator*{\argmax}{arg\,max}

\title{Is Encoder-Decoder Redundant for Neural Machine Translation?}

\author{
Yingbo Gao \qquad Christian Herold \quad Zijian Yang \qquad Hermann Ney \\
Human Language Technology and Pattern Recognition Group \\
Computer Science Department\\
RWTH Aachen University \\
D-52056 Aachen, Germany \\
{\tt \{ygao|herold|zyang|ney\}@cs.rwth-aachen.de}
}

\begin{document}

\maketitle
\begin{abstract}
Encoder-decoder architecture is widely adopted for sequence-to-sequence modeling tasks.
For machine translation, despite the evolution from long short-term memory networks to Transformer networks, plus the introduction and development of attention mechanism, encoder-decoder is still the de facto neural network architecture for state-of-the-art models.
While the motivation for decoding information from some hidden space is straightforward, the strict separation of the encoding and decoding steps into an encoder and a decoder in the model architecture is not necessarily a must.
Compared to the task of autoregressive language modeling in the target language, machine translation simply has an additional source sentence as context.
Given the fact that neural language models nowadays can already handle rather long contexts in the target language, it is natural to ask whether simply concatenating the source and target sentences and training a language model to do translation would work.
In this work, we investigate the aforementioned concept for machine translation.
Specifically, we experiment with bilingual translation, translation with additional target monolingual data, and multilingual translation.
In all cases, this alternative approach performs on par with the baseline encoder-decoder Transformer, suggesting that an encoder-decoder architecture might be redundant for neural machine translation.
\end{abstract}

\section{Introduction}

Sequence-to-sequence modeling is often approached with Neural Networks (NNs), prominently encoder-decoder NNs, nowadays.
For the task of Machine Translation (MT), which is by definition also a sequence-to-sequence task, the default choice of NN topology is also an encoder-decoder architecture.
For example, in early works like \citet{kalchbrenner2013recurrent}, the authors already make the distinction between their convolutional sentence model (encoder) and recurrent language model (decoder) conditioned on the former.
In follow-up works like \citet{sutskever2014sequence} and \citet{cho2014properties, cho2014learning}, the concept of encoder-decoder network is further developed.
While extensions such as attention \cite{bahdanau2014neural}, multi-task learning \cite{luong2015multi}, convolutional networks \cite{gehring2017convolutional} and self-attention \cite{vaswani2017attention} are considered for sequence-to-sequence learning, the idea of encoding information into some hidden space and decoding from that hidden representation sticks around.

Given the success and wide popularity of the Transformer network \cite{vaswani2017attention}, many works focus on understanding and improving individual components, e.g. positional encoding \cite{shaw-etal-2018-self}, multi-head attention \cite{voita-etal-2019-analyzing}, and an alignment interpretation of cross attention \cite{alkhouli-etal-2018-alignment}.
In works that go a bit further and make bigger changes in terms of modeling, e.g. performing round-trip translation \cite{tu2017neural} and going from autoregressive to non-autoregressive \cite{gu2017non}, the encoder-decoder setup itself is not really questioned.
In the mean time, it is not to say that the field is completely dominated by one approach.
Because works like the development of direct neural hidden Markov model \cite{wang2017hybrid, wang-etal-2018-neural-hidden, wang2021transformer}, investigation into dropping attention and separate encoding and decoding steps \cite{Press2018YouMN} and going completely encoder-free \cite{tang-etal-2019-understanding} do exist, where the default encoder-decoder regime is not directly applied.

Meanwhile, in the field of language modeling, significant progress is achieved with the wide application of NNs.
With the progress from early feedforward language models (LMs) \cite{bengio2000neural}, to the successful long short-term memory network LMs \cite{sundermeyer2012lstm}, and to the more recent Transformer LMs \cite{Irie2019LanguageMW}, the modeling capacity of LMs nowadays is much more than their historic counterparts.
This is especially true when considering some of the most recent extensions, such as large-scale modeling \cite{brown2020gpt3}, modeling very long context \cite{dai-etal-2019-transformer} and going from autoregressive modeling to non-autoregressive modeling \cite{devlin-etal-2019-bert}.
Because MT can be thought of as a contextualized language modeling task with the source sentence being additional context, one natural question is if simply concatenating the source and target sentences and train an LM to do translation would work \cite{irie2020:phd}.
This idea is simple and straightforward, but special care needs to be taken about the attention mechanism and source reconstruction.
In this work, we explore this alternative approach and conduct experiments in bilingual translation, translation with additional target monolingual data and multilingual translation.
Our results show that dropping the encoder-decoder architecture and simply treating the task of MT as contextualized language modeling is sufficient to obtain state-of-the-art results in translation.
This result has several subtleties and implications, which we discuss in Sec.\ref{sec:discussions}, and opens up possibilities for more general interfaces for multimodal modeling.

\section{Related Work}

In the literature, few but interesting works exist which closely relate to the idea mentioned above.
In \citet{mikolov2012context}, the authors mention the possibility to use source sentence as context for contextualized language modeling.
In \citet{he2018layer}, with the intuition to coordinate the learning of Transformer encoder and decoder layer by layer, the authors share the encoder and decoder parameters and learn a joint model on concatenated source and target sentences.
However, no explicit source side reconstruction loss is included.
Similarly, in \citet{irie2020:phd}, a small degradation in translation quality is observed when a causal mask is used and no source reconstruction is included.
Because the masking is critical for correctly modeling the dependencies regarding the concatenated sequence, in \citet{raffel2020exploring}, the authors put special focus on discussing the differences and implications of three types of attention masks.
In \citet{wang2021language}, the authors expand upon the idea and propose a two-step decaying learning rate schedule to reconstruct the source sentence to regularize the training process.
In that work, the authors show competitive performance compared to Transformer baselines in several settings.
More recently, in \citet{zhang2022examining}, the authors also use a language-modeling-style source side reconstruction loss to regularize the model, and additionally explore the model scaling cross-lingual transfer capabilities.
Another work that explores the long-context modeling potential of LMs is \citet{hawthorne2022general}, where data from domains other than translation is included in model training.
\citet{hao2022language} is a more recent addition to this direction of research, where LM as a general interface for multimodal data is investigated.
Because our focus is in MT, we refer to such a model, where encoder-decoder architecture is dropped and an LM is used to model the concatenation of source and target sentence, as Translation Language Models (TLMs\footnote{To be differentiated from TLMs in \citet{conneau2019cross}, where the pretraining objective is cloze task at both source and target side, using bilingual context.}).

The work by \citet{wang2021language} is probably the most directly related work compared to our work, therefore we believe it is important to highlight the similarities and differences between their work and ours.
The core concept of dropping encoder-decoder architecture is similar between \citet{wang2021language} and our work, and competitive performance of TLMs compared to encoder-decoder models in various settings is achieved in both works.
However, we additionally explore the task of autoencoding in the source side, adding Bidirectional-Encoder-Representations-from-Transformers-style (BERT) noise \cite{devlin-etal-2019-bert}, using alternative learning rate schedules, training MT models with back-translated (BT) data and doing multilingual training.
Further, we discuss subtleties and implications associated with the TLM.

\section{Methodology}

The core concept of TLM is to concatenate the source and the target sentences and treat the translation task as a language modeling task during training.
The two majors points of concern are the attention mechanism and the source-side reconstruction loss.
In this section, we explain the details related to these two points, and additionally discuss the implications when additional target-side monolingual data or multilingual data is available.

\subsection{Translation Language Model}

Denoting the source words/subwords as $f$ and the target words/subwords as $e$, with running indices $j$ in $J$ and $i$ in $I$ respectively, the usual way to approach the translation problem in encoder-decoder models is to directly model the posterior probabilities via a discriminative model $P(e_1^I|f_1^J)$.
This is used in the Transformer and can be expressed as:

\begin{equation*}
    P(e_1^I|f_1^J) = \prod_{i=1}^I P(e_i | e_0^{i-1}, f_1^J).
\end{equation*}

The model is usually trained with the cross entropy criterion (often regularized with label smoothing \cite{gao-etal-2020-towards}), and the search aims to find the target sentence $\hat{e}_1^{\hat{I}}$ with the highest probability (often approximated with beam search):

\begin{equation*}
    \begin{gathered}
        L_{\text{MT}} = - \sum_{i=1}^I \log P(e_i|e_0^{i-1}, f_1^J), \\
        \hat{e}_1^{\hat{I}} = \argmax_{e_1^I, I} \{ \log P(e_1^I|f_1^J) \}.
    \end{gathered}
\end{equation*}

Alternatively, one can model the joint probability of the source and target sentences via a generative model $P(f_1^J, e_1^I)$ and it can be expressed as:

\begin{equation*}
    P(f_1^J, e_1^I) = \prod_{j=1}^J P(f_j|f_0^{j-1}) \prod_{i=1}^I P(e_i | e_0^{i-1}, f_1^J).
\end{equation*}

Here, because $f_1^J$ is given at search time, and $\argmax_{e_1^I, I} P(f_1^J, e_1^I) = \argmax_{e_1^I, I} P(e_1^I|f_1^J)$, the search stays the same as in the baseline case.
But the training criterion has an additional loss term on the source sentence, which we refer to as reconstruction loss ($L_{\text{RE}}$), the learning rate $\lambda$ of which can be controlled by some schedule:

\begin{equation*}
    \begin{gathered}
        L_{\text{RE}} = - \sum_{j=1}^J \log P(f_j|f_0^{j-1}),\\
        L_{\text{TLM}} = \lambda L_{\text{RE}} +  L_{\text{MT}}.
    \end{gathered}
\end{equation*}

One can think of the reconstruction loss (decomposed in an autoregressive manner here, but it does not have to be) as a second task in addition to the translation task, or simply a regularization term for better learning of the source hidden representations.
Although this formulation is simple and straightforward, there could be variations in how the source side dependencies are defined.

\begin{figure}[ht]
     \centering
     \begin{subfigure}[b]{0.49\textwidth}
         \centering
         \vspace{0em}
         \includegraphics[trim={4em 0em 0em 0em},clip,width=\textwidth]{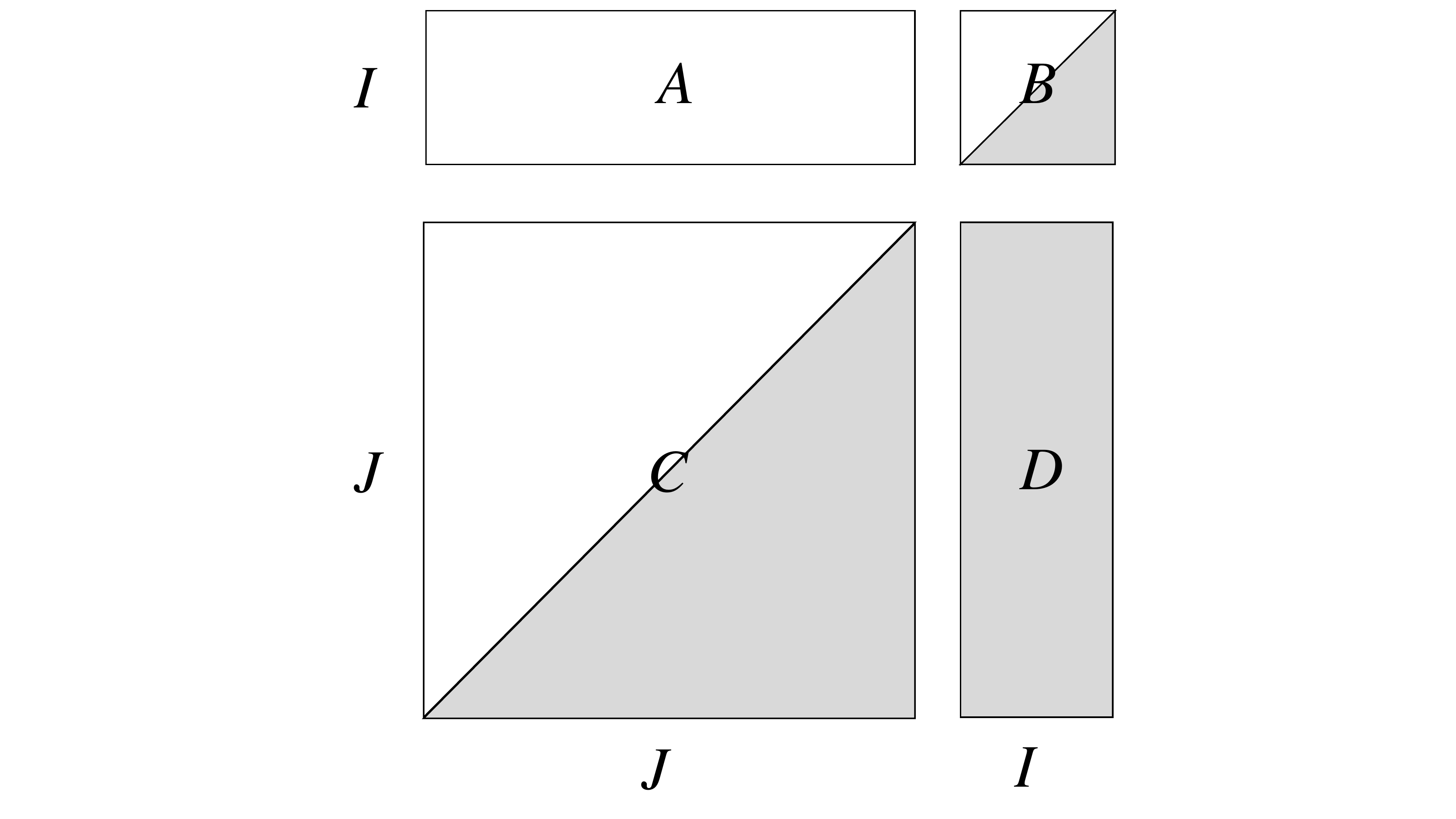}
         \caption{source-side triangular mask}\label{fig:mask1}
     \end{subfigure}
     \vfill
     \begin{subfigure}[b]{0.49\textwidth}
         \centering
         \vspace{2em}
         \includegraphics[trim={4em 0em 0em 0em},clip,width=\textwidth]{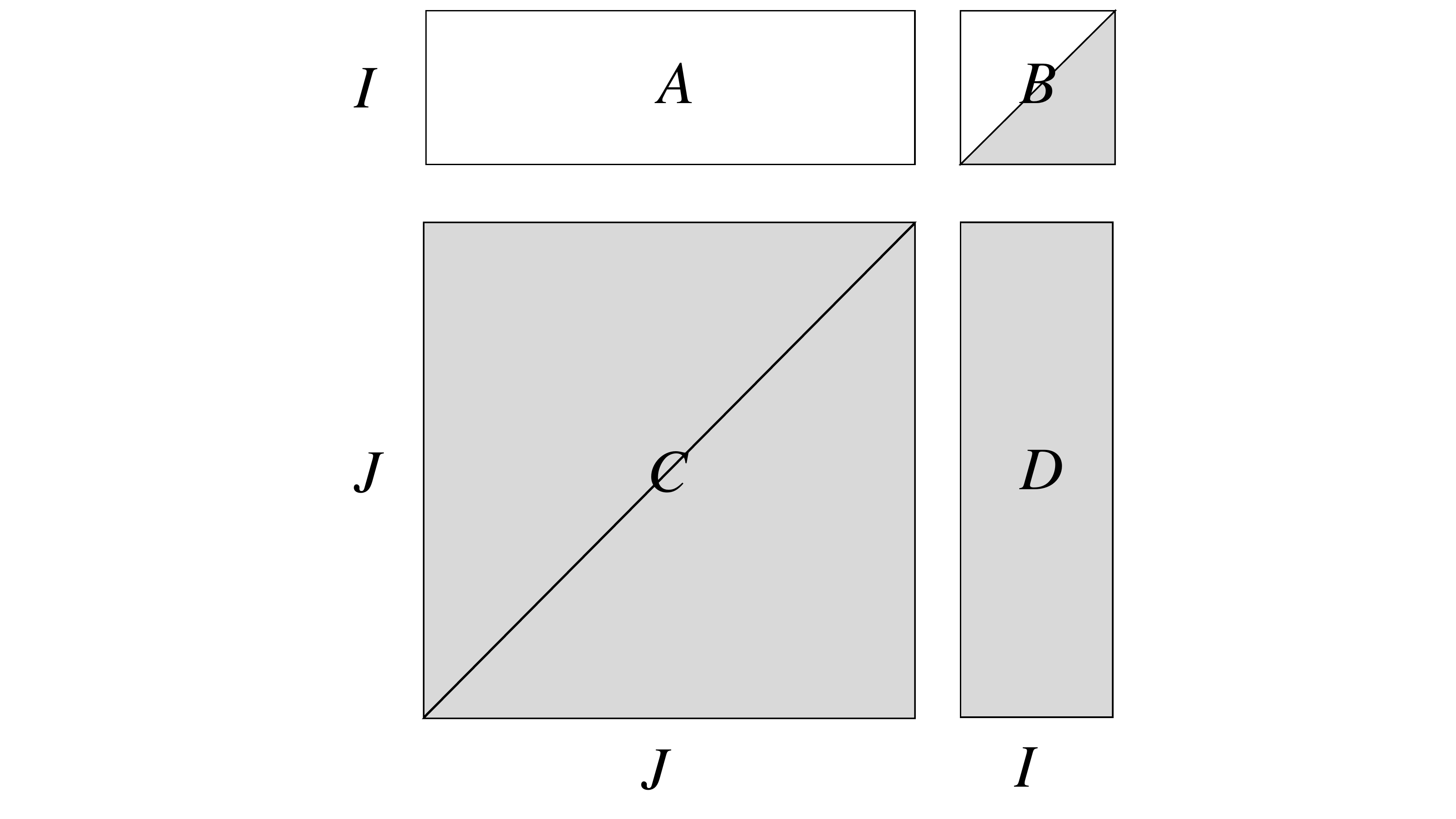}
         \caption{source-side full mask}\label{fig:mask2}
     \end{subfigure}
        \caption{Attention masks in TLM with (a) a triangular mask, and (b) a full mask, at the source side. The horizontal direction is the query direction and the vertical direction is the key direction. Shaded areas mean that the attention is valid and white areas mean that the attention is blocked. The matrices C, B, and D correspond to the encoder self attention, the decoder self attention and encoder-decoder cross attention in Transformer, respectively. The matrix A is whitened in both cases because we should not allow the source positions attend to future target positions.}
        \label{fig:masks}
\end{figure}

\begin{figure*}[ht]
     \centering
     \begin{subfigure}[b]{0.49\textwidth}
         \centering
         \includegraphics[width=\textwidth]{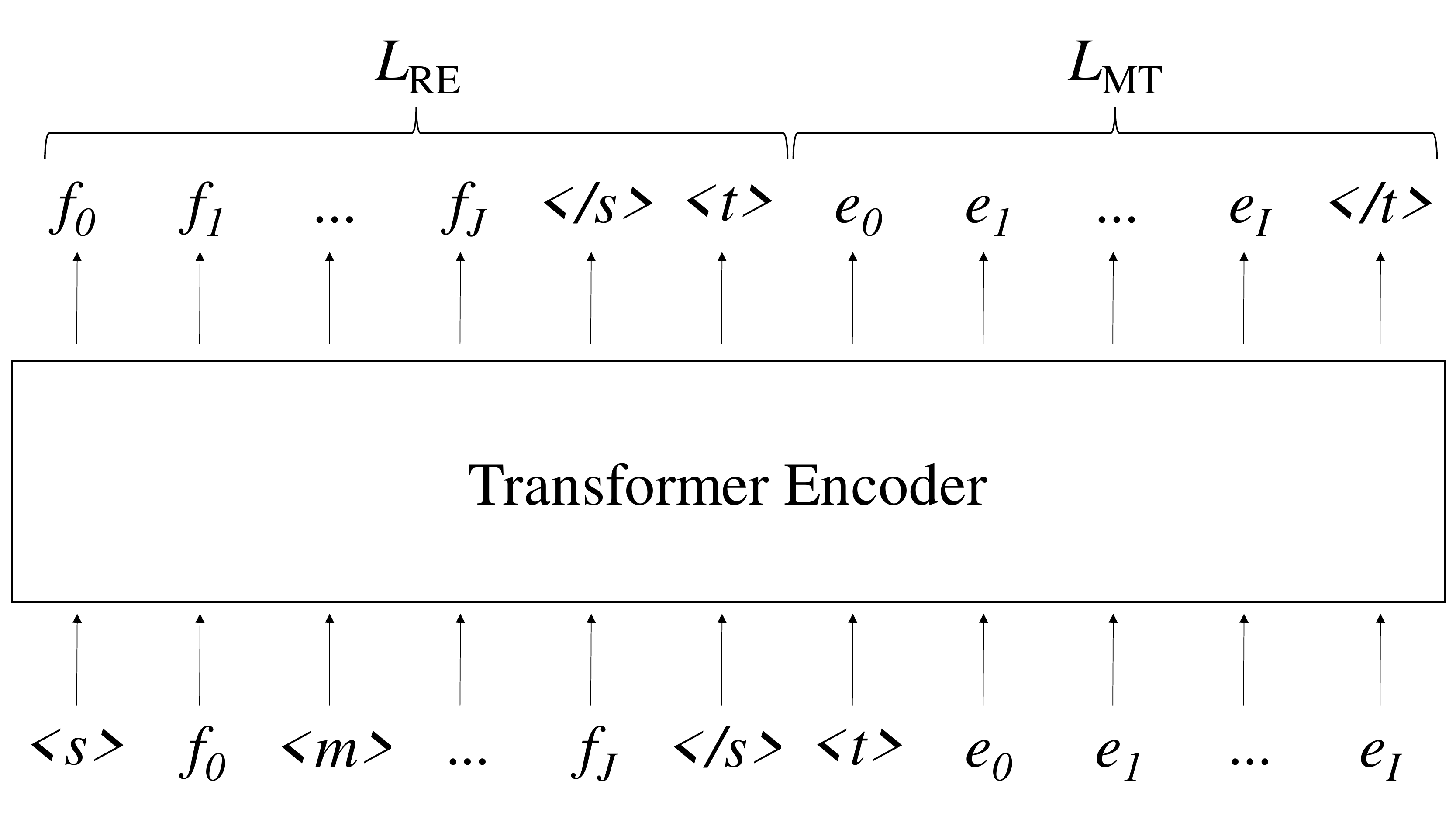}
         \caption{source-side shift in TLM}
         \label{fig:shift}
     \end{subfigure}
     \hfill
     \begin{subfigure}[b]{0.49\textwidth}
         \centering
         \includegraphics[width=\textwidth]{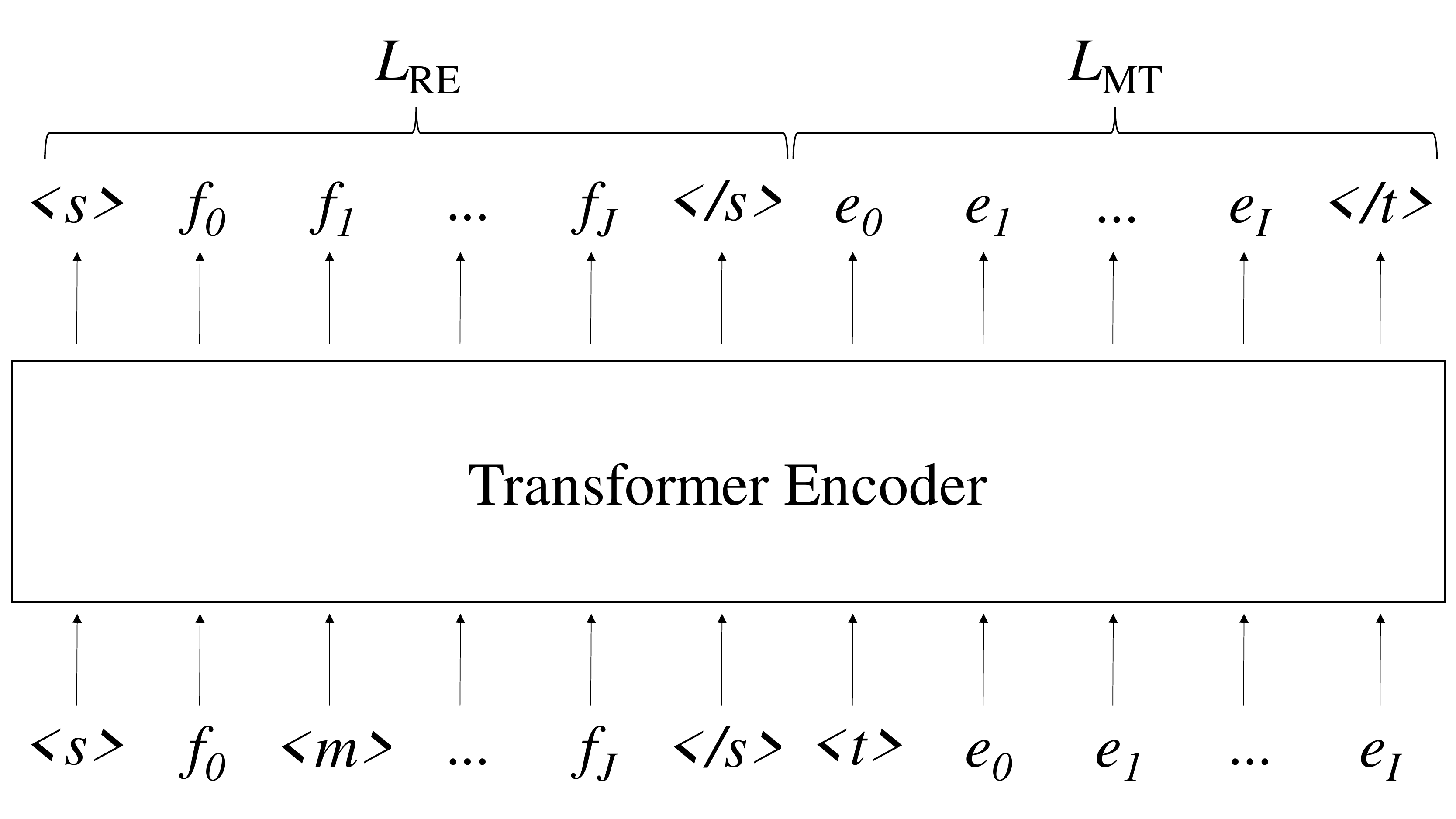}
         \caption{no source-side shift in TLM}
         \label{fig:noshift}
     \end{subfigure}
        \caption{Shifting versus no shifting of the output at the source side in TLM. The output at the target side is shifted in both cases. <$s$>, <$/s$>, <$t$> and <$/t$> are artificial start and end of sentence symbols at the source and target side respectively\footnotemark. <$m$> denotes BERT-style \cite{devlin-etal-2019-bert} randomly masked tokens. When matrix C in Fig.\ref{fig:masks} is triangular, (a) corresponds to a language modeling objective. When C is full, (b) corresponds to an auto-encoding objective. During search, <$s$>, $f_0$, ..., $f_J$, <$/s$>, <$t$> is presented to the model, and beam search is done by minimizing $L_{\text{MT}}$.}
        \label{fig:shifting}
\end{figure*}

\subsubsection{On the Attention Mechanism}

In the original Transformer \cite{vaswani2017attention} model, the attention mechanism is used in three places, namely, a $J \times J$ encoder self attention matrix, a $I \times I$ decoder self attention matrix and a $J \times I$ encoder-decoder cross attention matrix.
As shown in Fig.\ref{fig:masks}, they correspond to matrices C, B and D respectively.
The attention masks in B and D are straightforward.
The triangular attention mask in the B matrix needs to be causal by definition, because otherwise target positions may attend to future positions and cheat.
The attention mask in D needs to be full, because we want each target position to be able to look at each source position so that there is no information loss.
However, the attention mask in C is how some of the previous works differ.
For example, a triangular attention mask like in Fig.\ref{fig:mask1} is used in \citet{irie2020:phd}, while a full attention mask like in Fig.\ref{fig:mask2} is used in \citet{he2018layer}.
\citet{raffel2020exploring} and \citet{zhang2022examining} also discuss the differences in masking patterns in the matrix C similar to what we do here.
\citet{wang2021language} do not make clear what type of attention masks is used in C in their paper, and we do not find a public repository associated with their paper to further investigate it.

\interfootnotelinepenalty=10000
\footnotetext{The exact format of the tags is not important so long as uniquely identifiable translation direction tags are used, be it source and target tags like <$s$>, <$/s$>, <$t$>, <$/t$>, or direction tags like <$s2t$>, or even only the target language tag <$t$>. While the later two reduce the total sequence length, the former is more versatile when data from multiple languages or multiple modalities is considered.}

In our case, we consider both the triangular and full attention mask patterns for C, because both have good intuitions.
The triangular mask is closer to the original objective of learning the joint distribution $P(f_1^J, e_1^I)$, while the full mask enables better information flow because early source positions also have access to future source positions to come up with better hidden representations.
That said, later we show through experiments, that for the task of MT, it is clearly better to use a full attention mask for C in TLM.

The matrix A in Fig.\ref{fig:masks} is whitened throughout this work, because we do not allow the source positions attend to target positions.
However, theoretically, when decoding position $i$, one could allow all source positions $1, 2, ..., J$ to attend to all previous target positions $1, 2, ..., i-1$.
This can be done by using a $(J+I) \times (J+I) \times I$ attention mask tensor.
The extended $I$ dimension is target-position-dependent, providing a different view of the $(J+I) \times (J+I)$ matrix for each target position.
Intuitively, this has the potential to serve as an implicit fertility model.

\subsubsection{On the Reconstruction Loss}\label{sec:reconstruction}

In the paper by \citet{wang2021language}, the source side reconstruction is formulated as an autoregressive language modeling task.
However, that does not have to be the case.
For example, one can make the distinction to shift or not shift the output at the source side, as shown in Fig.\ref{fig:shifting}.
When the source output is shifted, $L_{\text{RE}}$ is a normal language modeling cross entropy loss.
When the source output is not shifted, $L_{\text{RE}}$ is an auto-encoding loss.
Additionally considering the matrix C in Fig.\ref{fig:masks}, assuming no source input noise is introduced, then when C is full, or when C is triangular but the source output is not shifted, the source-side reconstruction becomes a trivial copying task.

On top of the reconstruction loss formulation, one can also apply noises to the source side input.
This can be viewed as a regularization or a data augmentation trick, such that the source side information is corrupted to a certain degree to help the generalization ability of the model.
In this work, we consider the BERT-style \cite{devlin-etal-2019-bert} noises, where 15\% of source positions are picked at random, and 80\%, 10\% and 10\% of the tokens in this positions are replaced with <$m$>, a random token or unchanged, respectively.
Different to the BERT paper though, in addition to the cloze task in the masked positions, we also keep the cross entropy losses in the unmasked positions.
One can of course go over the 15\% \cite{wettig2022should} limit or apply softer noises \cite{gao-etal-2019-soft, gao2020unifying}, but we do not further expand in this direction because it is beyond our initial goal to verify the necessity of the encoder-decoder architecture.

One more thing that can be tuned for the reconstruction loss is the learning rate schedule.
In \citet{wang2021language}, a two-step linear decaying function is used, where $\lambda$ linearly decays to 0.1 until a certain number of gradient update steps $\tau$, and decays with a smaller rate after $\tau$.
Here, we additionally consider schedules where the learning rate $\lambda$: (a) is constant at zero, (b) is constant at one, (c) two-step linearly decays like in \citet{wang2021language} and (d) decays exponentially as $\lambda_t = \exp ( - \ln 0.1 t / \tau )$.
Similar to (c), the schedule (d) decays to 0.1 at gradient update step $\tau$ as well.

\subsection{Bilingual and Monolingual Training}

For MT, target-side monolingual data is often available in large quantities and is shown to be helpful for the main task of translation when used in one way or another \cite{koehn-etal-2007-moses, wuebker-etal-2012-jane, freitag-etal-2014-jane, sennrich-etal-2016-improving, gulcehre2017integrating, domhan-hieber-2017-using, stahlberg-etal-2018-simple, edunov-etal-2018-understanding, graca-etal-2019-generalizing}.
Broadly speaking, they can be categorized into three approaches: 1. ensembling with an external language model, 2. multi-task training with additional language modeling objective and 3. training with back-translated data with artificial source and true target.
Evidence so far is that back-translation works the best among the three \cite{wmt-2021-machine}.

For TLM, these three approaches are all applicable, but with implications.
First, ensembling is not very relevant because of the additional training and storage requirements, and also it is against the philosophy of TLM where we want to make the encoder-decoder model more compact.
Second, the multi-task training is interesting because while some previous work have dedicated layers to perform the language modeling task \cite{gulcehre2015using, gulcehre2017on}, such multi-task training on TLM actually trains all model parameters in the auxiliary language modeling task.
Third, the back-translation approach is worth looking at because it delivers the best results in encoder-decoder models so far and experiments comparing TLM with the baseline under this setting are necessary to justify whether or not we can throw away the encoder-decoder architecture.

\subsection{Multilingual Training}

Another important setting where TLM needs to be compared to the baseline encoder-decoder model is when multilingual data is used in training.
Broadly speaking, multilingual models can refer to systems that translate in one-to-many, many-to-one, many-to-many, or even source-to-target and target-to-source manners.
The major benefits of training multilingual models \cite{johnson2017google, aharoni-etal-2019-massively} are: more compact models via shared parameters and transfer/zero-shot learning capabilities due to inherit similarities in some languages.
While there exist works that propose to use language-specific sub-networks to take into consideration the parameter capacity needed for each language, e.g. in \citet{lin-etal-2021-learning}, it is more common to simply train one joint model where the model parameters are shared across all languages.

For TLM, the task of multilingual training is straightforwards as well.
One can simply concatenate each translation pair into one longer sequence, add corresponding translation direction tags, and feed the concatenated sequence to the TLM model.
In other words, all the hidden parameters of the model can be shared across all translation directions, and one simply needs to pay attention to the word embeddings such that words/subwords/tokens from different languages are mapped into the same embedding space for further processing, similar to what is done for encoder-decoder models.

\section{Experiments}

To verify the performance of TLM compared to the baseline encoder-decoder Transformer model, we perform experiments on four machine translation datasets.
Specifically, we experiment with the International Conference on Spoken Language Translation (IWSLT) \cite{iwslt-2014} 2014 German-to-English (de-en), the Conference on Machine Translation (WMT) 2016 English-to-Romanian (en-ro) \cite{bojar-etal-2016-findings}, 2019 Chinese-to-English (zh-en) \cite{barrault-etal-2019-findings} datasets.
Additionally, for multilingual experiments, we create a custom multilingual (multi.) dataset from news-commentary v16 \cite{tiedemann-2012-parallel}, performing translation among three languages, German (de), Spanish (es), and French (fr), in six direction: de-es, es-de, de-fr, fr-de, es-fr, fr-es\footnote{The data is retrieved from \url{https://data.statmt.org/news-commentary/v16/}. We take all bilingual data in the six directions. For the raw data in each direction, we take the first 3000 lines as test data and the last 3000 lines as development data.}.
For the monolingual data, we sample 5M sentences from the English News crawl monolingual corpus\footnote{\url{https://data.statmt.org/news-crawl/}}.
To create synthetic zh-en data, we employ our en-zh Transformer model to do back-translation \cite{sennrich-etal-2016-improving}.
The data is pre-processed with the Byte Pair Encoding (BPE) \cite{sennrich-etal-2016-neural} algorithm.
We lowercase the text for de-en and for the other language pairs, we leave the original casing as is.
The statistics of the datasets are summarized in Tab.\ref{tab:stats}.

\begin{table}[ht]
\centering
\begin{tabular}{crrr}
\hline
dataset & vocab. & train pairs & test pairs \\ \hline
de-en & 10k & 0.2M & 6k \\
en-ro & 20k & 0.6M & 2k \\
multi. & 32k & 1.7M & 18k \\
zh-en & 47k & 17.0M & 4k \\ \hline
\end{tabular}
\caption{\label{tab:stats} Statistics of the datasets.}
\end{table}

We implement the Transformer model and the TLM model with different options such as using different attention masks, shifting versus not shifting the source output, adding or not adding BERT-style \cite{devlin-etal-2019-bert} noises and different learning rate schedules, in PyTorch \cite{paszke2019pytorch}.
The back-translation and multilingual experiments are done by adding corresponding language tags to the concatenation of source and target sentences.

We follow the training and search hyperparameters as closely as possible to the original Transformer \cite{vaswani2017attention} paper.
Note that, when searching with TLM, the entire source sentence until (and including) the target start of sentence <$t$> is fed into the NN.
The beam search is then carried out only on the target outputs.
We report translation performances in \textsc{Bleu} \cite{papineni-etal-2002-bleu} and \textsc{Ter} \cite{snover-etal-2006-study} scores using the \texttt{MultEval} tool from \citet{clark-etal-2011-better}.

\subsection{An Encoder-Only Model}\label{sec:grid_search}

First, we consider the necessity of encoder-decoder architecture by comparing our encoder-only TLM with the baseline Transformer model, on de-en and en-ro.
Essentially, we perform a grid search over four hyper-parameters regarding the source reconstruction:

\begin{enumerate}\setlength\itemsep{-0.2em}
    \item Language modeling (shifting source output, Fig.\ref{fig:shift}) versus autoencoding (not shifting source output, Fig.\ref{fig:noshift}).
    \item Triangular (see Fig.\ref{fig:mask1}) versus full (Fig.\ref{fig:mask2}) attention mask.
    \item No source input noise versus BERT-style \cite{devlin-etal-2019-bert} source input noise (Sec.\ref{sec:reconstruction}).
    \item Constant learning rate $\lambda$ for $L_{\text{RE}}$ at zero or one, or the two-step linear \cite{wang2021language} or exponential decay (Sec.\ref{sec:reconstruction}).
\end{enumerate}

Due to the limited length, we only highlight the interesting points from our observations and append the full grid-search table (Tab.\ref{tab:grid}) in the appendix for the interested reader.
For the discussions below, we consider one hyperparameter each time and pick the best set of other hyperparameters from the grid search, to take into considerations of possible correlations among different hyperparameters.

\subsubsection{Both Autoencoding and Language Modeling Work}

First, we see that both shifting and not shifting the source output at the source side seem to work for TLM.
As shown in Tab.\ref{tab:shift?}, when picking the best set of other hyperparameters, TLMs trained with either of the auxiliary task can perform on par with the encoder-decoder baseline within $\pm 0.2\%$ absolute \textsc{Bleu} score fluctuations.

\begin{table}[ht]
\centering
\begin{tabular}{cccc}
\hline
arch. & task & de-en & en-ro \\ \hline
enc-dec & - & 34.9 & 26.0 \\ \hline
\multirow{2}{*}{enc-only} & LM & 34.7 & 26.2 \\
 & AE & 35.0 & 26.0 \\ \hline
\end{tabular}
\caption{\label{tab:shift?} \textsc{Bleu} scores of language modeling (LM) versus autoencoding (AE) at the source side.}
\end{table}

\subsubsection{Full Attention Over Source Is Necessary}

\begin{table}[ht]
\centering
\begin{tabular}{cccc}
\hline
arch. & mask & de-en & en-ro \\ \hline
enc-dec & - & 34.9 & 26.0 \\ \hline
\multirow{2}{*}{enc-only} & triangular & 34.4 & 25.6 \\
 & full & 35.0 & 26.2 \\ \hline
\end{tabular}
\caption{\label{tab:mask?} \textsc{Bleu} scores of triangular versus full attention mask at the source side.}
\end{table}

Looking at the source attention mask (Tab.\ref{tab:mask?}), it is clear that a triangular leads to degradation in translation performance.
One interesting setup is when the attention mask is triangular but the task is autoencoding, i.e. no shift in source outputs.
One may argue that the model is allowed to cheat on the auxiliary task $L_{\text{RE}}$ because the diagonals in the attention mask is not masked out, however, for the translation task, it is possible that source hidden representations learned from being able to look at future source positions is more beneficial.

\subsubsection{BERT-Style Noise Is Slightly Helpful}

Moving on to the source-side noise, adding BERT-style \cite{devlin-etal-2019-bert} seems to slightly boost the translation performance.
This observation agrees with past experiences where augmenting the training data with artificial noise regularizes the model for better generalization \cite{hill-etal-2016-learning, kim-etal-2018-improving, kim-etal-2019-effective, gao-etal-2019-soft, gao2020unifying}.

\begin{table}[ht]
\centering
\begin{tabular}{cccc}
\hline
arch. & noise & de-en & en-ro \\ \hline
enc-dec & - & 34.9 & 26.0 \\ \hline
\multirow{2}{*}{enc-only} & none & 34.6 & 26.1 \\
 & BERT & 35.0 & 26.2 \\ \hline
\end{tabular}
\caption{\label{tab:noise?} \textsc{Bleu} scores with and without BERT-style \cite{devlin-etal-2019-bert} noises at the source side.}
\end{table}

\subsubsection{Loss Schedule Is Not Critical}

Contrary to \citet{wang2021language} and also to our surprise, the learning rate schedule for $\lambda$ does not seem to be critical for obtaining good translation performance with TLM.
As shown in Tab.\ref{tab:schedule?}, even without the reconstruction loss $L_{\text{RE}}$, i.e. when $\lambda$ is constant at zero, the \textsc{Bleu} score of the TLM is still comparable with the baseline transformer.
Of course one needs to tune the other hyperparameters, it is still interesting that the model is able to learn decent source hidden representations even without any auxiliary training signal.

\begin{table}[ht]
\centering
\begin{tabular}{cccc}
\hline
arch. & schedule & de-en & en-ro \\ \hline
enc-dec & - & 34.9 & 26.0 \\ \hline
\multirow{4}{*}{enc-only} & 0 & 34.9 & 26.2 \\
 & 1 & 34.5 & 26.0 \\
 & lin & 34.7 & 25.8 \\
 & exp & 34.8 & 26.1 \\ \hline
\end{tabular}
\caption{\label{tab:schedule?} \textsc{Bleu} scores with different learning rate schedules of $\lambda$. "lin" refers to the two-step learning rate decay in \citet{wang2021language} and "exp" refers to the exponential decay introduced in Sec.\ref{sec:reconstruction}.}
\end{table}

\subsubsection{Parameter Count Needs to Be the Same}

Although the hyperparameters mentioned so far have different degrees of influence on the final \textsc{Bleu} score, one hyperparameter that governs the overall performance of TLM is the total learnable parameter count.
Similar to \citet{wang2021language}, the encoder-only model needs to have a similar amount of parameters to reach the performance of the Transformer baseline.
Here, we vary the number of Transformer encoder layers in TLM and compare with the baseline Transformer to illustrate this point.
An autoencoding loss is used without shifting the source outputs, noises are added to the source inputs, and a fixed $\lambda=1$ is used for the encoder-only TLMs in Tab.\ref{tab:params?}.
It can be seen that, when the TLM is under- or over- parametrized, underfitting and overfitting happens respectively, leading to worse performances.

\begin{table}[ht]
\centering
\begin{tabular}{ccccc}
\hline
\multirow{2}{*}{arch.} & \multirow{2}{*}{\#layers} & \multirow{2}{*}{\#params} & \multicolumn{2}{c}{de-en} \\
 &  &  & \textsc{Bleu} & \textsc{Ter} \\ \hline
enc-dec & 6-6 & 36.9M & 34.9 & 44.5 \\ \hline
\multirow{4}{*}{enc-only} & 5 & 15.9M & 33.5 & 46.2 \\
 & 10 & 26.4M & 34.9 & 44.6 \\
 & 15 & 36.9M & 35.0 & 44.7 \\
 & 20 & 47.4M & 34.8 & 45.1 \\ \hline
\end{tabular}
\caption{\label{tab:params?} \textsc{Bleu} and \textsc{Ter} scores of models of different sizes. For the encoder-decoder model, 6-6 means 6 encoder layers and 6 decoder layers.}
\end{table}

\begin{table}[h]
\centering
\begin{tabular}{lccc}
\hline
\multicolumn{1}{c}{\multirow{2}{*}{arch.}} & \multirow{2}{*}{devPPL} & \multicolumn{2}{c}{zh-en} \\
\multicolumn{1}{c}{} &  & \textsc{Bleu} & \textsc{Ter} \\ \hline
enc-dec & 6.91 & 23.2 & 60.5 \\
\; + back-translation & 6.21 & 24.6 & 59.4 \\ \hline
enc-only & 6.90 & 23.1 & 60.5 \\
\; + LM & 6.70 & 23.0 & 61.4 \\
\; + back-translation & 6.18 & 24.7 & 59.4 \\ \hline
\end{tabular}
\caption{Transformer versus TLM, with and without additional monolingual target side data.}
\label{tab:mono}
\end{table}

\begin{table*}[ht]
\centering
\begin{tabular}{ccccccccc}
\hline
arch. & devPPL & de-es & es-de & de-fr & fr-de & es-fr & fr-es & overall \\ \hline
enc-dec & 6.17 & 25.7 & 19.1 & 21.3 & 16.9 & 24.6 & 26.2 & 22.5 \\
enc-only & 6.06 & 25.5 & 18.8 & 20.7 & 16.6 & 24.4 & 26.0 & 22.3 \\ \hline
\end{tabular}
\caption{\label{tab:multi}\textsc{Bleu} scores of multilingual translation with encoder-decoder Transformer and encoder-only TLM. Here, we train both the encoder-decoder baseline model as well as the encoder-only TLM until the same number of steps and pick the best checkpoint according to the best development set perplexity. The overall score is calculated over the concatenation of the test sets and is not the average of the previous columns.}
\end{table*}

\subsection{Bilingual and Monolingual Training}

The streamlined architecture of TLM allows us to easily include monolingual data during training, without the need to create synthetic parallel data and without having to modify the architecture in any way.
The system is simply trained jointly on the translation and language modeling tasks.
We compare this training strategy to the most common way of including monolingual data in MT training, namely back-translation and experiment on the high resource zh-en task.
The results are shown in Tab.\ref{tab:mono}.

As expected, the additional synthetic data from back-translation leads to an improvement in both, development set perplexity (devPPL) and translation quality, for the Transformer and TLM.
Including the monolingual data directly in the TLM does also improve perplexity, but does not improve overall translation quality.

\vspace{1em}
\subsection{Multilingual Training}

The experimental results for the multilingual translation are summarized in Tab.\ref{tab:multi}.
Although the encoder-only TLM actually delivers better devPPL than the encoder-decoder Transformer baseline, the \textsc{Bleu} scores are slightly worse (about $-0.2\%$ absolute \textsc{Bleu}) across the board.
This mismatch between the development set perplexity and the test \textsc{Bleu} in NMT is also reported in previous work \cite{gao-etal-2020-towards}.
We believe this small difference is within acceptable noise range and conclude that the TLM is also on par with the baseline encoder-decoder model in multilingual translation.

\vspace{1em}
\section{Discussions}\label{sec:discussions}

Through extensive experiments, we show that the encoder-decoder architecture is not a must to achieve decent translation performance, because an encoder-only TLM is also capable of obtaining comparable performance when carefully tuned.
Here, we touch upon several important implications and subtleties that come with using TLMs.

First, although the encoder-decoder architecture is dropped, the cross attention is still existent in the TLM.
As shown in Fig.\ref{fig:masks}, the difference compared to the baseline is that for each target position $i$, the softmax needs to normalize the attention weights over $J+i$ instead of $J$.
However, because we know the softmax is decent at zeroing out certain positions, e.g. see Fig.1 in \citet{alkhouli-etal-2018-alignment}, this should not be a problem.
Next, although we do not expand on search in this paper, our internal experiments verify that the search with TLM behaves similarly to the baseline.
Further, one may wonder how separate source and target vocabularies should be handled in case of TLMs.
Here, we note that having separate source and target word embedding matrices is the same as concatenating them in the vocabulary size dimension into a bigger word embedding matrix for TLM.
What could pose as a problem is the increased length of the concatenated sequence.
This puts extra requirements to the model and its capabilities to model long context dependencies.
Note that, concatenation may not be the only way to combine the source and target contexts.
For instance, in the eager model proposed in \citet{Press2018YouMN}, the authors essentially "stack" instead of "concatenate".
Moreover, when decoding efficiency is critical, TLM may suffer because a separate decoder is not existent and each translation query goes through the entire network.
Another limitation is that the source side reconstruction loss considered in this work may also be applied to the Transformer baseline, and might change the picture when comparing the two.
That said, TLMs are undoubtedly exciting models opening new possibilities.
For example, with such generative models, generation of synthetic translation pairs from scratch can be easily done.
Another worth-to-mention application is end-to-end speech translation (ST).
While previous work, e.g. in \citet{bahar2021tight}, connects the encoder of the automatic speech recognition model and the decoder of the MT model, effectively throwing away 50\% of the pre-trained model parameters, TLMs can retain all pre-trained parameters and result in more compact end-to-end ST models.

\vspace{1em}
\section{Conclusion}

In this work, we question the long-standing encoder-decoder architecture for neural machine translation.
Through extensive experiments in various translation directions, considering back-translation and multilingual translation, we find that an encoder-only model can perform as good as an encoder-decoder model.
We further discuss implications and subtleties of such models to motivate further research into more compact models and more general neural network interfaces.

\section*{Acknowledgements}
This work was partially supported by the project HYKIST funded by the German Federal Ministry of Health on the basis of a decision of the German Federal Parliament (Bundestag) under funding ID ZMVI1-2520DAT04A, and by NeuroSys which, as part of the initiative “Clusters4Future”, is funded by the Federal Ministry of Education and Research BMBF (03ZU1106DA).

\bibliography{ref}
\bibliographystyle{acl}

\clearpage
\begin{appendices}

\onecolumn
\section{Grid Search Over Source Reconstruction Settings}
\begin{table*}[ht]
\centering
\begin{tabular}{|c|cccc|cc|cc|}
\hline
\multirow{2}{*}{architecture} & \multicolumn{4}{c|}{source reconstruction variant} & \multicolumn{2}{c|}{IWSLT14 de-en} & \multicolumn{2}{c|}{WMT16 en-ro} \\ \cline{2-9} 
 & \multicolumn{1}{c|}{task} & \multicolumn{1}{c|}{mask} & \multicolumn{1}{c|}{noise} & schedule & \multicolumn{1}{c|}{\textsc{Bleu}} & \textsc{Ter} & \multicolumn{1}{c|}{\textsc{Bleu}} & \textsc{Ter} \\ \hline
encoder-decoder & \multicolumn{1}{c|}{-} & \multicolumn{1}{c|}{-} & \multicolumn{1}{c|}{-} & - & \multicolumn{1}{c|}{34.9} & 44.5 & \multicolumn{1}{c|}{26.0} & 54.8 \\ \hline
\multirow{32}{*}{encoder-only} & \multicolumn{1}{c|}{\multirow{16}{*}{LM}} & \multicolumn{1}{c|}{\multirow{8}{*}{triangular}} & \multicolumn{1}{c|}{\multirow{4}{*}{none}} & 0 & \multicolumn{1}{c|}{33.5} & 46.0 & \multicolumn{1}{c|}{25.4} & 55.5 \\ \cline{5-9} 
 & \multicolumn{1}{c|}{} & \multicolumn{1}{c|}{} & \multicolumn{1}{c|}{} & 1 & \multicolumn{1}{c|}{34.4} & 45.3 & \multicolumn{1}{c|}{25.2} & 55.7 \\ \cline{5-9} 
 & \multicolumn{1}{c|}{} & \multicolumn{1}{c|}{} & \multicolumn{1}{c|}{} & lin & \multicolumn{1}{c|}{34.2} & 45.2 & \multicolumn{1}{c|}{25.5} & 55.4 \\ \cline{5-9} 
 & \multicolumn{1}{c|}{} & \multicolumn{1}{c|}{} & \multicolumn{1}{c|}{} & exp & \multicolumn{1}{c|}{34.6} & 45.2 & \multicolumn{1}{c|}{25.3} & 55.7 \\ \cline{4-9} 
 & \multicolumn{1}{c|}{} & \multicolumn{1}{c|}{} & \multicolumn{1}{c|}{\multirow{4}{*}{BERT}} & 0 & \multicolumn{1}{c|}{33.6} & 45.7 & \multicolumn{1}{c|}{25.4} & 55.6 \\ \cline{5-9} 
 & \multicolumn{1}{c|}{} & \multicolumn{1}{c|}{} & \multicolumn{1}{c|}{} & 1 & \multicolumn{1}{c|}{34.4} & 45.1 & \multicolumn{1}{c|}{25.2} & 55.8 \\ \cline{5-9} 
 & \multicolumn{1}{c|}{} & \multicolumn{1}{c|}{} & \multicolumn{1}{c|}{} & lin & \multicolumn{1}{c|}{34.4} & 45.4 & \multicolumn{1}{c|}{25.6} & 55.5 \\ \cline{5-9} 
 & \multicolumn{1}{c|}{} & \multicolumn{1}{c|}{} & \multicolumn{1}{c|}{} & exp & \multicolumn{1}{c|}{34.2} & 45.8 & \multicolumn{1}{c|}{25.4} & 55.3 \\ \cline{3-9} 
 & \multicolumn{1}{c|}{} & \multicolumn{1}{c|}{\multirow{8}{*}{full}} & \multicolumn{1}{c|}{\multirow{4}{*}{none}} & 0 & \multicolumn{1}{c|}{34.5} & 44.9 & \multicolumn{1}{c|}{25.8} & 55.4 \\ \cline{5-9} 
 & \multicolumn{1}{c|}{} & \multicolumn{1}{c|}{} & \multicolumn{1}{c|}{} & 1 & \multicolumn{1}{c|}{34.5} & 44.8 & \multicolumn{1}{c|}{25.9} & 55.0 \\ \cline{5-9} 
 & \multicolumn{1}{c|}{} & \multicolumn{1}{c|}{} & \multicolumn{1}{c|}{} & lin & \multicolumn{1}{c|}{34.5} & 44.9 & \multicolumn{1}{c|}{25.7} & 55.3 \\ \cline{5-9} 
 & \multicolumn{1}{c|}{} & \multicolumn{1}{c|}{} & \multicolumn{1}{c|}{} & exp & \multicolumn{1}{c|}{34.4} & 44.8 & \multicolumn{1}{c|}{26.1} & 54.8 \\ \cline{4-9} 
 & \multicolumn{1}{c|}{} & \multicolumn{1}{c|}{} & \multicolumn{1}{c|}{\multirow{4}{*}{BERT}} & 0 & \multicolumn{1}{c|}{34.5} & 45.1 & \multicolumn{1}{c|}{26.2} & 54.8 \\ \cline{5-9} 
 & \multicolumn{1}{c|}{} & \multicolumn{1}{c|}{} & \multicolumn{1}{c|}{} & 1 & \multicolumn{1}{c|}{34.4} & 44.9 & \multicolumn{1}{c|}{25.6} & 55.3 \\ \cline{5-9} 
 & \multicolumn{1}{c|}{} & \multicolumn{1}{c|}{} & \multicolumn{1}{c|}{} & lin & \multicolumn{1}{c|}{34.7} & 44.5 & \multicolumn{1}{c|}{25.8} & 55.3 \\ \cline{5-9} 
 & \multicolumn{1}{c|}{} & \multicolumn{1}{c|}{} & \multicolumn{1}{c|}{} & exp & \multicolumn{1}{c|}{34.6} & 44.9 & \multicolumn{1}{c|}{25.9} & 54.9 \\ \cline{2-9} 
 & \multicolumn{1}{c|}{\multirow{16}{*}{AE}} & \multicolumn{1}{c|}{\multirow{8}{*}{triangular}} & \multicolumn{1}{c|}{\multirow{4}{*}{none}} & 0 & \multicolumn{1}{c|}{32.2} & 47.2 & \multicolumn{1}{c|}{25.3} & 55.6 \\ \cline{5-9} 
 & \multicolumn{1}{c|}{} & \multicolumn{1}{c|}{} & \multicolumn{1}{c|}{} & 1 & \multicolumn{1}{c|}{32.5} & 46.2 & \multicolumn{1}{c|}{24.9} & 55.9 \\ \cline{5-9} 
 & \multicolumn{1}{c|}{} & \multicolumn{1}{c|}{} & \multicolumn{1}{c|}{} & lin & \multicolumn{1}{c|}{32.0} & 46.3 & \multicolumn{1}{c|}{25.2} & 55.8 \\ \cline{5-9} 
 & \multicolumn{1}{c|}{} & \multicolumn{1}{c|}{} & \multicolumn{1}{c|}{} & exp & \multicolumn{1}{c|}{32.0} & 46.8 & \multicolumn{1}{c|}{25.3} & 55.3 \\ \cline{4-9} 
 & \multicolumn{1}{c|}{} & \multicolumn{1}{c|}{} & \multicolumn{1}{c|}{\multirow{4}{*}{BERT}} & 0 & \multicolumn{1}{c|}{30.8} & 47.9 & \multicolumn{1}{c|}{25.1} & 56.1 \\ \cline{5-9} 
 & \multicolumn{1}{c|}{} & \multicolumn{1}{c|}{} & \multicolumn{1}{c|}{} & 1 & \multicolumn{1}{c|}{33.5} & 45.9 & \multicolumn{1}{c|}{25.1} & 56.0 \\ \cline{5-9} 
 & \multicolumn{1}{c|}{} & \multicolumn{1}{c|}{} & \multicolumn{1}{c|}{} & lin & \multicolumn{1}{c|}{31.5} & 47.5 & \multicolumn{1}{c|}{25.2} & 55.7 \\ \cline{5-9} 
 & \multicolumn{1}{c|}{} & \multicolumn{1}{c|}{} & \multicolumn{1}{c|}{} & exp & \multicolumn{1}{c|}{33.6} & 45.9 & \multicolumn{1}{c|}{25.5} & 55.6 \\ \cline{3-9} 
 & \multicolumn{1}{c|}{} & \multicolumn{1}{c|}{\multirow{8}{*}{full}} & \multicolumn{1}{c|}{\multirow{4}{*}{none}} & 0 & \multicolumn{1}{c|}{34.4} & 45.1 & \multicolumn{1}{c|}{25.8} & 55.2 \\ \cline{5-9} 
 & \multicolumn{1}{c|}{} & \multicolumn{1}{c|}{} & \multicolumn{1}{c|}{} & 1 & \multicolumn{1}{c|}{34.0} & 45.3 & \multicolumn{1}{c|}{25.9} & 55.1 \\ \cline{5-9} 
 & \multicolumn{1}{c|}{} & \multicolumn{1}{c|}{} & \multicolumn{1}{c|}{} & lin & \multicolumn{1}{c|}{33.8} & 45.7 & \multicolumn{1}{c|}{25.7} & 55.3 \\ \cline{5-9} 
 & \multicolumn{1}{c|}{} & \multicolumn{1}{c|}{} & \multicolumn{1}{c|}{} & exp & \multicolumn{1}{c|}{34.0} & 45.5 & \multicolumn{1}{c|}{25.7} & 55.3 \\ \cline{4-9} 
 & \multicolumn{1}{c|}{} & \multicolumn{1}{c|}{} & \multicolumn{1}{c|}{\multirow{4}{*}{BERT}} & 0 & \multicolumn{1}{c|}{34.9} & 45.0 & \multicolumn{1}{c|}{25.8} & 55.3 \\ \cline{5-9} 
 & \multicolumn{1}{c|}{} & \multicolumn{1}{c|}{} & \multicolumn{1}{c|}{} & 1 & \multicolumn{1}{c|}{35.0} & 45.0 & \multicolumn{1}{c|}{26.0} & 55.1 \\ \cline{5-9} 
 & \multicolumn{1}{c|}{} & \multicolumn{1}{c|}{} & \multicolumn{1}{c|}{} & lin & \multicolumn{1}{c|}{34.7} & 45.0 & \multicolumn{1}{c|}{25.7} & 55.4 \\ \cline{5-9} 
 & \multicolumn{1}{c|}{} & \multicolumn{1}{c|}{} & \multicolumn{1}{c|}{} & exp & \multicolumn{1}{c|}{34.8} & 44.8 & \multicolumn{1}{c|}{25.8} & 55.4 \\ \hline
\end{tabular}
\caption{\label{tab:grid}Grid search of four source-reconstruction-related hyperparameters on de-en and en-ro. LM means to shift the source-side outputs and the auxiliary task corresponds to autoregressive language modeling, and AE means to not shift and corresponds to an autoencoding task. Our interpretations of the table are given in Sec.\ref{sec:grid_search}}
\end{table*}

\end{appendices}

\end{document}